\documentclass[runningheads]{llncs}

\usepackage{eccv}

\usepackage{eccvabbrv}

\usepackage{xcolor}
\usepackage{graphicx}
\usepackage{booktabs}
\usepackage{subcaption}
\usepackage{multirow}
\usepackage{makecell}
\usepackage{arydshln}
\usepackage{amsmath,amssymb}
\usepackage[table]{xcolor}

\usepackage[accsupp]{axessibility}  

\usepackage[]{hyperref}

\begin{document}

\title{Multimodal Floorplan Encoding: Learning Dense Modality-Invariant Representations}

\titlerunning{Multimodal Floorplan Encoding}

\author{Xavier Anadón\inst{1}\thanks{Corresponding author: xanadon@unizar.es} \and
Rémi Pautrat\inst{2} \and
Rui Wang\inst{2}}

\authorrunning{X. Anadón et al.}

\institute{University of Zaragoza, Spain \and Microsoft, Switzerland}

\maketitle

\begin{abstract}
Floorplans arise in many forms, from vector CAD drawings to raster renderings and sensor-derived density maps. This heterogeneity makes it difficult to build learning systems that transfer across modalities and support geometry-centric tasks such as alignment and retrieval.
We introduce the \emph{Multimodal Floorplan Encoder} (MMFE), which maps diverse 2D indoor representations into a shared dense latent grid.
MMFE combines a frozen DINOv3 backbone with a trainable Dense Prediction Transformer (DPT) head, and is trained with a per-cell Information Noise-Contrastive Estimation (InfoNCE) objective that aligns spatially corresponding regions across modalities while using all other cells as negatives.
To improve robustness to geometric distortions, we incorporate controlled similarity transformations and enforce geometric consistency through feature-grid warping.
On Structured3D, a held-out out-of-domain dataset, MMFE improves cross-modal dense matching, enables robust similarity alignment with RANSAC, and yields strong retrieval when paired with learned aggregation.
\end{abstract}

\section{Introduction}
\label{sec:intro}

Two-dimensional floorplans are among the most widespread and long-lived
representations of indoor space: compact, human-readable, cheap to store, and
largely unaffected by the alterations that break appearance-based maps (e.g., changes in furniture, light, or paint).

In robotics,
architectural plans and BIM models serve as global references for localization
and mapping, by matching geometric primitives~\cite{zimmerman2022semantic} or
higher-level topological structures~\cite{shaheer2023isgraphs}; by directly localizing a camera inside a
floorplan~\cite{howardjenkins2021lalaloc,lalalocpp,min2022laser}; or by reasoning about
the depth of the surrounding structure~\cite{f3loc,semanticrays,unloc}. Beyond localization, floorplans are the
working representation for architectural analysis, layout generation and
retrieval~\cite{nauata2020housegan,hu2020graph2plan}, and the target output of
indoor reconstruction pipelines~\cite{liu2018floornet,heat,roomformer}.

What makes this heterogeneous landscape hard to unify is that "a floorplan" is
not a single kind of data. The same apartment may be available as a vector CAD drawing, as a
raster plan with furniture, hatching and text annotations, as a stylized
rendering of wall and opening polygons, or as a sensor-derived observation. Consequently, most learned floorplan systems are trained for one
modality and one dataset, and degrade once the rendering style
or the sensing process changes. Generic visual encoders offer only partial
relief: as we show in \cref{sec:experiments}, DINOv2~\cite{dinov2} / DINOv3~\cite{dinov3} features of two
modalities of the \emph{same} floorplan are dominated by appearance rather than
by layout. What is missing is a representation of floorplans that is agnostic to
how the floorplan was produced.

We address this gap with the \emph{Multimodal Floorplan Encoder} (MMFE), which
maps any of these indoor representations into a single, shared latent grid.
MMFE is deliberately \emph{dense}: rather than collapsing a floorplan into one
global vector as in contrastive multimodal pretraining
(e.g.\ CLIP~\cite{radford2021clip}), it predicts a grid of local
descriptors, so that spatial correspondence is preserved and geometry-centric
tasks become directly accessible. It pairs a frozen DINOv3
backbone~\cite{dinov3} with a trainable Dense Prediction Transformer (DPT)
head~\cite{dpt}, trained with a per-cell InfoNCE objective~\cite{oord2018infomax}.
The resulting descriptors are modality-invariant enough to be used in downstream tasks such as multimodal alignment and retrieval. We validate these tasks on a held-out generalization test set, where MMFE substantially outperforms generic dense features. The individual components we build on are deliberately standard; our contribution lies in the multimodal floorplan formulation and the training recipe rather than in a new architecture.

\paragraph{Contributions.}
\begin{itemize}
    \item A \textbf{multimodal formulation of floorplan representation learning}, in
    which heterogeneous 2D representations of the same scene are treated as views to
    be bound together in a single dense latent grid, preserving geometric structure
    instead of collapsing each floorplan into a global embedding.
    \item \textbf{MMFE}, an instantiation of this formulation from standard
    components --- a frozen DINOv3 backbone, a DPT head, and a per-cell InfoNCE
    objective in which positives are spatially corresponding cells of two different
    modalities and the remaining cells of the batch act as a large set of hard
    negatives.
    \item A \textbf{per-cell contrastive objective} that aligns corresponding regions across modalities while leveraging a large set of hard negatives from the remaining cells in the batch.
    \item A \textbf{geometric robustness training recipe} based on controlled similarity perturbations and feature-grid warping, validated out-of-domain on cross-modal \textbf{alignment} and \textbf{retrieval}, where MMFE substantially outperforms strong generic dense-feature baselines.
\end{itemize}

\section{Related Work}
\label{sec:related_work}

\paragraph{Floorplan representations.}
Architectural and CAD pipelines represent floorplans with polylines and polygons
for walls, doors and windows, optionally enriched with semantics as in BIM
models~\cite{shaheer2023isgraphs,standfest2022swiss,ZInD}.
However, floorplans are also encountered as raster images~\cite{kalervo2019cubicasa5kdatasetimprovedmultitask}.
A complementary line of work \emph{generates} floorplans from observations: raster-to-vector approaches parse existing drawings into
structured geometry~\cite{liu2017rastertovector,kalervo2019cubicasa5kdatasetimprovedmultitask},
while reconstruction methods recover layouts from 3D scans or images~\cite{liu2018floornet, heat, roomformer}.
When only sensor data is available, the floorplan is a
projection of it: an orthographic point-density map of a 3D
reconstruction~\cite{avetisyan2024scenescript,Structured3D} or a sparse set of 2D range
returns~\cite{zimmerman2022semantic}. These formats describe the \emph{same geometry}
under very different appearance statistics, which is precisely the
heterogeneity MMFE is designed to absorb.

\paragraph{Floorplan-based localization and matching.}
Classical approaches match geometric primitives extracted from range data to the
plan~\cite{zimmerman2022semantic}, and iS-Graphs~\cite{shaheer2023isgraphs}
lifts this to hierarchical graph matching. Learned methods increasingly localize from RGB
alone: LaLaLoc~\cite{howardjenkins2021lalaloc} and
LaLaLoc++~\cite{lalalocpp} embed layouts and panoramic observations in a shared
latent space; LASER~\cite{min2022laser} renders the plan as a point set and
matches pose-conditioned features; and
F$^3$Loc~\cite{f3loc} with its semantic extension~\cite{semanticrays}
exploit predicted structural depth inside a filtering framework, while
UnLoc~\cite{unloc} leverages off-the-shelf monocular depth models. Closer to matching, FloorplanNet~\cite{floorplannet} aligns 3D submaps to
2D plans via topometric graphs. All of these systems are engineered around one particular
sensing-to-map pairing; our goal is orthogonal, namely a single dense
representation in which \emph{any} pair of floorplan modalities becomes directly
comparable.

\paragraph{Multimodal representation learning.}
Aligning heterogeneous inputs in a shared embedding space is the central idea
behind aligning image–text pairs~\cite{radford2021clip, zhai2023siglip}, while 
ImageBind~\cite{girdhar2023imagebind} extends the principle to align six modalities. In the 3D indoor
domain, CrossOver~\cite{sarkar2025crossover} aligns images, point clouds,
CAD models and text at the scene level. These methods produce a \emph{global} embedding per
input, discarding the spatial layout. MMFE keeps the multimodal
contrastive formulation but moves the alignment down to the level of individual
spatial cells, treating floorplan modalities as the views to be bound together.

\paragraph{Contrastive learning and dense representations.}
Contrastive and metric-learning objectives shape
embedding spaces through pairwise supervision. From the original contrastive ~\cite{hadsell2006dimensionality} and triplet ~\cite{schroff2015facenet}
losses to formulations that exploit more of the batch ~\cite{ohsong2016lifted, sohn2016npair, wang2019multisimilarity}. Among these, InfoNCE~\cite{oord2018infomax}
has become the dominant choice for representation learning, underpinning approaches like
SimCLR~\cite{chen2020simclr} and MoCo~\cite{he2020moco}. While global objectives
discard spatial information, dense variants contrast local descriptors instead~\cite{wang2021densecl, xie2021pixpro}. We adopt a dense InfoNCE formulation, but in a
\emph{multimodal} setting: positives are not two augmentations of one image but
spatially corresponding cells of two different floorplan modalities of the same
scene.

\section{Multimodal Floorplan Dataset}
\label{sec:dataset}

\paragraph{Floorplan Modalities:}

\begin{figure}[t]
    \centering
    \begin{subfigure}[b]{0.45\linewidth}
        \centering
        \includegraphics[width=\linewidth]{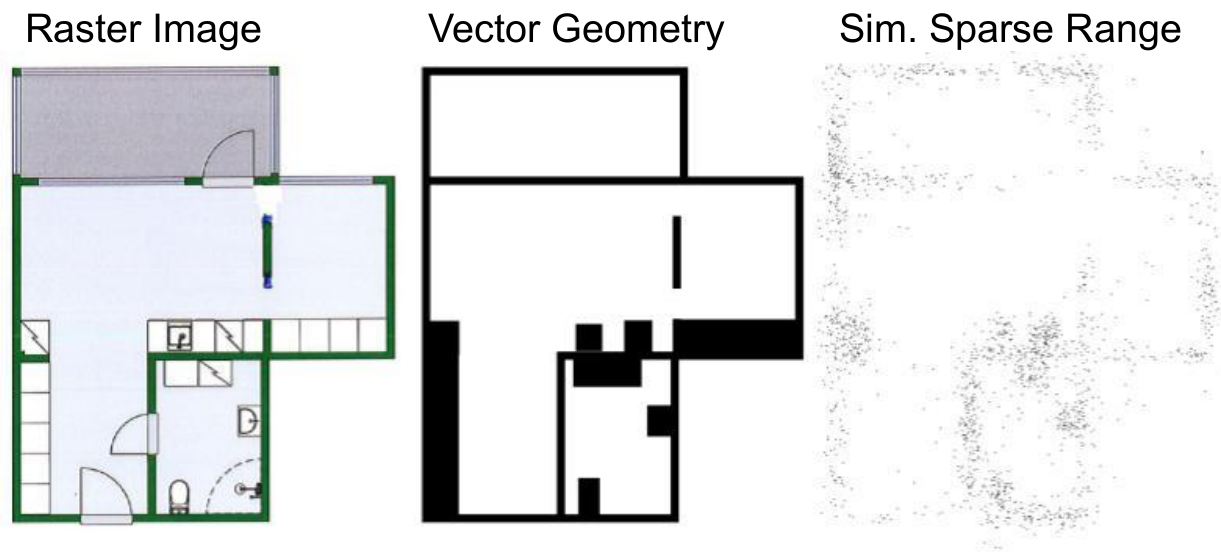}
        \caption{CubiCasa5K \cite{kalervo2019cubicasa5kdatasetimprovedmultitask}}
    \end{subfigure}
    \hfill
    \begin{subfigure}[b]{0.45\linewidth}
        \centering
        \includegraphics[width=\linewidth]{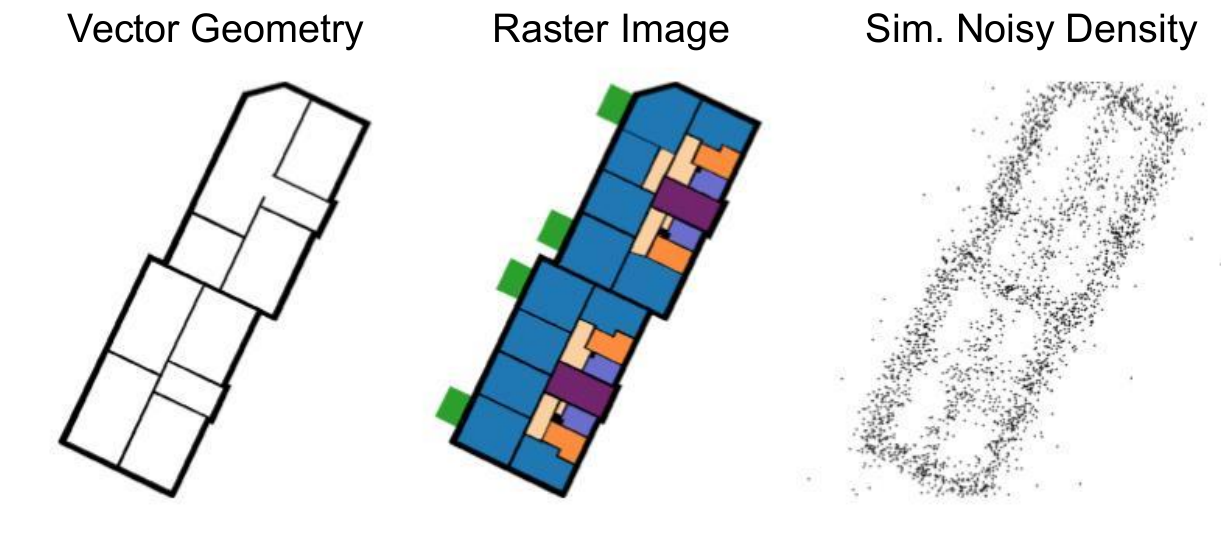}
        \caption{Swiss Dwellings \cite{standfest2022swiss}}
    \end{subfigure}

    \vspace{0.5em}

    \begin{subfigure}[b]{0.45\linewidth}
        \centering
        \includegraphics[width=\linewidth]{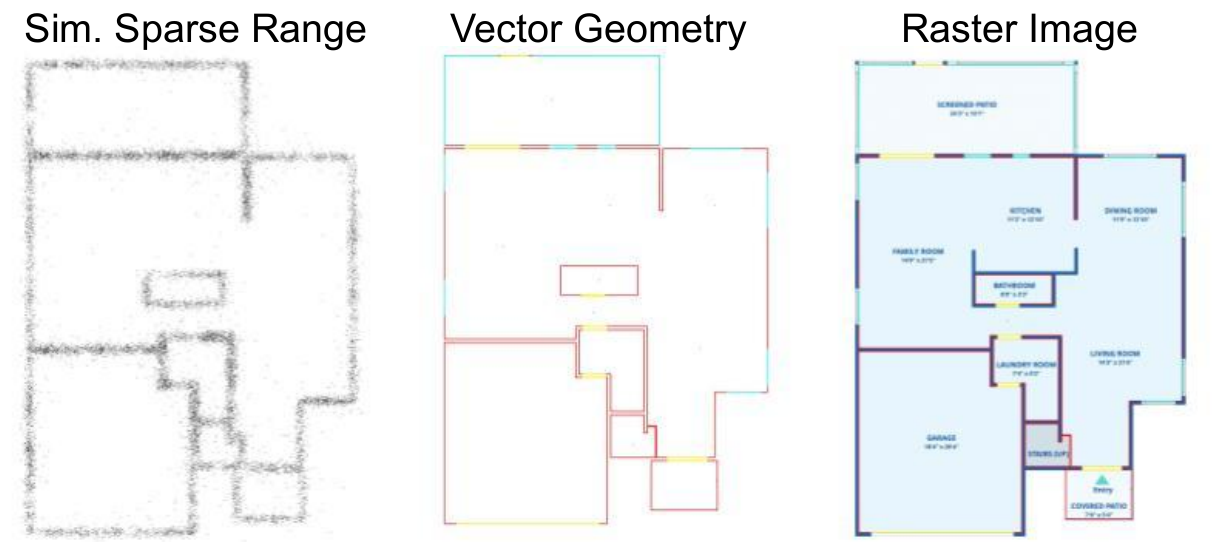}
        \caption{ZInD \cite{ZInD}}
    \end{subfigure}
    \hfill
    \begin{subfigure}[b]{0.45\linewidth}
        \centering
        \includegraphics[width=\linewidth]{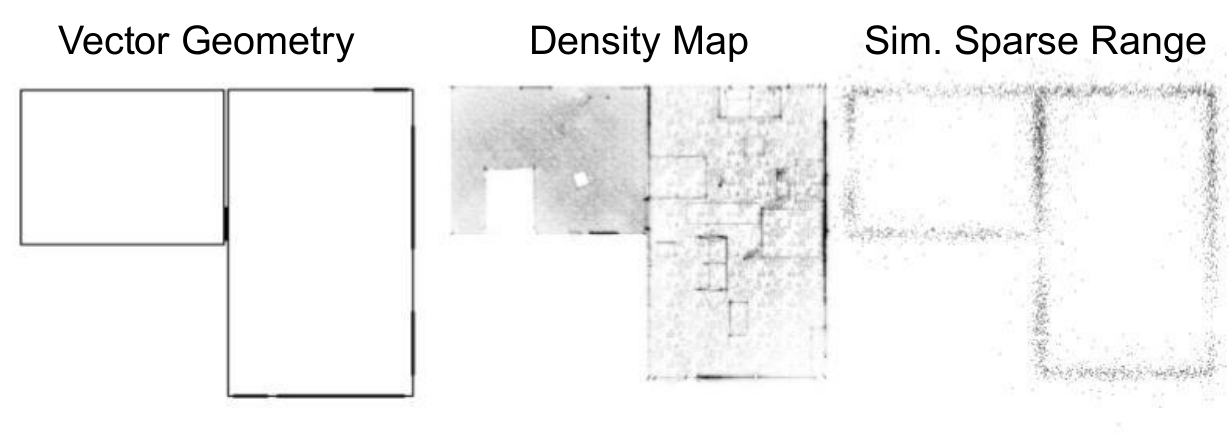}
        \caption{Aria SE \cite{avetisyan2024scenescript}}
    \end{subfigure}

    \vspace{0.5em}

    \begin{subfigure}[b]{0.55\linewidth}
        \centering
        \includegraphics[width=\linewidth]{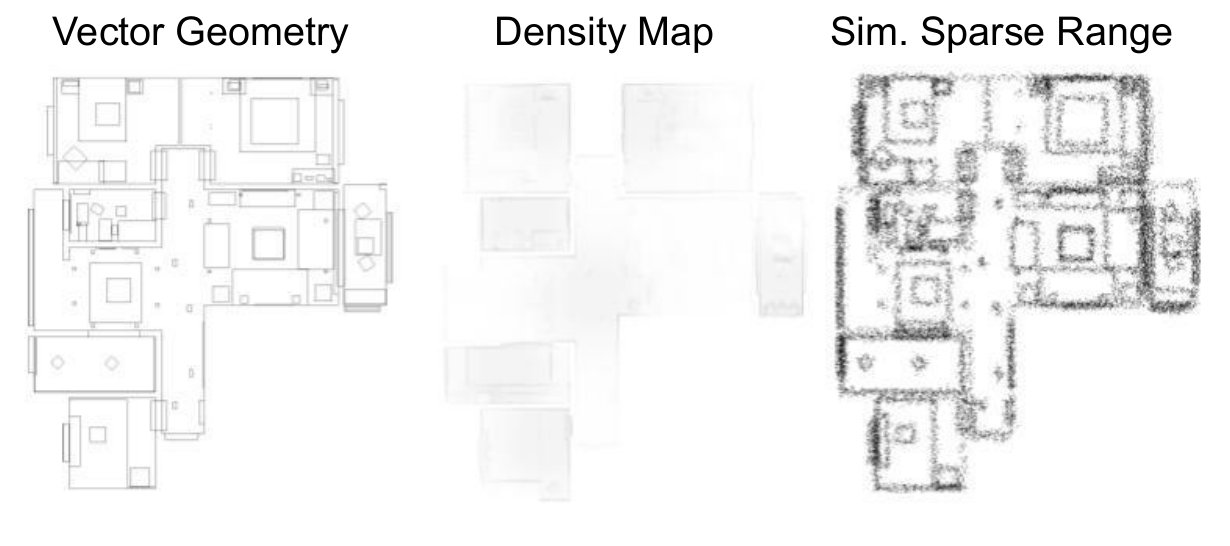}
        \caption{Structured3D (Test) \cite{Structured3D}}
    \end{subfigure}
    \caption{Example floorplan sources used for training (a - d) and evaluation (e). }
    \label{fig:datasets}
\end{figure}

Before describing the encoder, we make explicit the input space it is expected
to absorb. We group the indoor representations considered in this work into
four families, illustrated in \cref{fig:datasets}:

\begin{itemize}
    \item \textbf{Vector Geometry.} Walls, doors and windows are
    available as vector geometry and are rasterized into clean line drawings. Appearance is
    minimal and fully determined by the rendering convention (line thickness,
    which element classes are drawn), so two renderings of the same layout can
    still differ substantially in style.
    \item \textbf{Raster Image.} The floorplan is given as an
    image produced for human consumption. Besides the
    layout, it can contain furniture symbols, hatching patterns, dimension lines, and text.
    \item \textbf{Density Maps.} A 3D representation of the scene is
    projected orthographically onto the ground plane and rasterized into a
    density map.
    \item \textbf{Sparse Range Scans.} A small number of 2D range
    returns is available instead of a dense map, yielding a partial, noisy
    outline of the structure.
\end{itemize}

Throughout the paper, a \emph{modality pair} denotes two such representations of the same floorplan, and all supervision is derived from pairs of this kind.
\Cref{fig:datasets} illustrates representative samples. We held out the Structured3D~\cite{Structured3D} dataset from the training sets to test out-of-domain generalization. All results in this paper are reported on the validation split of this dataset.

In total, our training split contains $14{,}939$ floorplans across four sources (\cref{tab:datasets-training}), which amounts to a total of $44{,}817$ modality pairs. These floorplans are at apartment or single-floor scale rather than building scale: for reference, Structured3D contains $21{,}835$ rooms across $3{,}500$ house models~\cite{Structured3D} and Swiss Dwellings $242{,}257$ rooms across $42{,}207$ apartments~\cite{standfest2022swiss}, i.e. approximately six rooms per plan in both cases. Layouts with hundreds of interconnected rooms are outside the scope of our current setup (\cref{sec:limitations}).

\paragraph{Synthetic 2D-LiDAR modality.}
As we are not aware of any dataset that pairs \textit{Sparse Range Scans} with the sources above at a comparable number of examples, we additionally simulate a sparse 2D LiDAR-like observation from the binary layout mask $M\in\{0,1\}^{H\times W}$ of each floorplan. The goal is not to create a physically accurate sensor model, but rather a modality that captures the partial and noisy structural cues characteristic of planar range measurements. We first build a smooth, spatially varying density field by low-pass filtering random noise and applying a non-linear contrast transform, then normalize it over the layout interior to obtain a sampling distribution $P(p)$. Given a target density $\rho\in(0,1]$, we then draw $N=\lfloor \rho\,|\{p:M(p)=1\}|\rfloor$ points without replacement according to $P$, which controls the sparsity of the observation. Each sampled point is perturbed by Gaussian noise $\varepsilon\sim\mathcal{N}(0,\sigma^2 I)$ in pixel units, with a small fraction assigned a larger variance to emulate outliers, and the points inside the image boundaries are rasterized into a sparse density map. Consequently, this modality does not perfectly model viewpoint-dependent visibility, occlusions, or sensor-specific artifacts; we discuss the implications in \cref{sec:limitations}.

\begin{table}[]
\centering
\caption{Datasets and splits used in this work. In bold we highlight the most important numbers: the total training set, the in-domain validation set used only for design choices, and the out-of-domain validation set. Out-of-domain training set (gray) is never seen during training.}
\setlength{\tabcolsep}{6pt}
\begin{tabular}{l l c c }
\hline
\textbf{Dataset Type} & \textbf{Dataset Name} & \textbf{Training} & \textbf{Validation} \\
\hline
\multirow{5}{*}{Training Datasets}
 & CubiCasa5K \cite{kalervo2019cubicasa5kdatasetimprovedmultitask}        & 4{,}200  & 200  \\
 & Swiss Dwellings \cite{standfest2022swiss}  & 4{,}572  & 800  \\
 & Zillow Indoor (ZInD) \cite{ZInD} & 2{,}167  & 278   \\
 & Aria SE \cite{avetisyan2024scenescript}          & 4{,}000  & 1{,}000   \\
 & \textbf{Total}    & \textbf{14{,}939} & \textbf{2{,}278}  \\
\hline
Evaluation Dataset
 & Structured3D  \cite{Structured3D}    & \textcolor{gray!90}{3{,}000}  & \textbf{247}  \\
\hline
\end{tabular}
\label{tab:datasets-training}
\end{table}

\section{Method}
\label{sec:method}

\subsection{Dense Multimodal Encoder}
\begin{figure}[t]
    \centering
    \includegraphics[width=\linewidth]{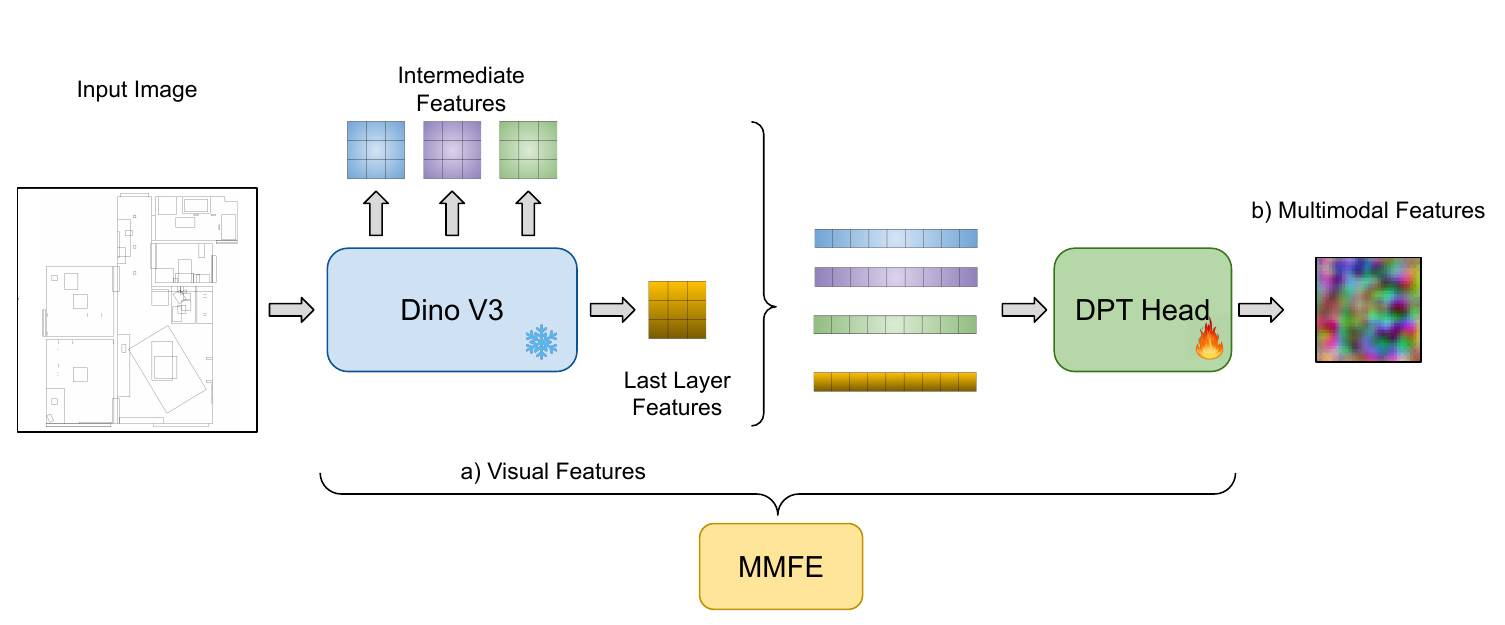}
    \caption{MMFE architecture: a frozen DINOv3 backbone~\cite{dinov3} with a trainable DPT-style head~\cite{dpt} producing a $32\times32\times32$ latent grid.}
    \label{fig:arch}
\end{figure}

Given an input floorplan image $x$ (generated from any modality), MMFE produces a latent tensor
\[
F = f_{\theta}(x) \in \mathbb{R}^{H \times W \times D},
\]
with $H=W=32$ and descriptor dimension $D=32$. The spatial domain is aligned with the original image, but downsampled to a lower resolution to reduce computations in downstream tasks.
The encoder combines (i) a frozen DINOv3 backbone~\cite{dinov3} and (ii) a trainable Dense Prediction Transformer (DPT) ~\cite{dpt} head inspired by MoGe~\cite{moge1} that fuses multi-scale transformer features into a structured grid (\cref{fig:arch}). We justify the selection of this architecture in \cref{sec:experiments}.

\subsection{Per-Cell Multimodal InfoNCE}
Training uses paired modalities $(x, x')$ of the same scene, producing $F=f_\theta(x)$ and $F'=f_\theta(x')$.
Both modalities are rendered into a common canonical floorplan frame at the same resolution, so cell $(u,v)$ in $F$ and in $F'$ refers to the same scene location; this defines the ground-truth cell correspondence used for supervision.
This shared frame follows directly from how the modalities are produced, so no explicit registration step is required.
For scenes reconstructed from 3D data, the annotations that yield the rendered floorplan and the projected point clouds are already aligned in 3D, and we preserve this alignment when projecting to 2D. 
The synthetic point-density maps inherit the frame of the modality from which they are generated.
Therefore, corresponding cells initially share the same coordinates by construction.
For each cell $(u,v)$, we treat the corresponding cell in the other modality as the positive,

\[
z = F[u,v], \qquad z^+ = F'[u,v],
\]
and all other cells in the batch (including same-image cells) as negatives.
We optimize a per-cell InfoNCE loss~\cite{oord2018infomax}:
\begin{equation}
\mathcal{L}_{u,v} = -\log
\frac{\exp(\mathrm{sim}(z,z^+)/\tau)}
{\exp(\mathrm{sim}(z,z^+)/\tau) + \sum_{n \in \mathcal{N}_{u,v}} \exp(\mathrm{sim}(z,n)/\tau)}.
\label{eq:cell_infonce}
\end{equation}
where $\mathrm{sim}(\cdot,\cdot)$ denotes the cosine similarity between $\ell_2$-normalized feature vectors, and $\tau$ is the temperature parameter that controls the sharpness of the softmax distribution.
Note that spatially adjacent cells of the same scene are also included among the negatives $\mathcal{N}_{u,v}$; since neighboring cells are often near-duplicates, this acts as a strong hard-negative signal that pushes the encoder toward spatially discriminative descriptors without the need of more explicit negative mining.

\subsection{Geometric Robustness via Similarity Perturbations}
\label{sec:transforms}

Downstream alignment requires robustness to planar transformations, so we
perturb the training pairs with elements of the planar similarity group
$\mathrm{Sim}(2)$. We sample transforms from a family
parameterized by three \emph{boundary parameters} representing rotation angle, translation and scale change
$(\theta_{\max}, t_{\max}, \Delta s)$, drawing
\begin{equation}
    \theta \sim \mathcal{U}(-\theta_{\max}, \theta_{\max}), \quad
    t_x, t_y \sim \mathcal{U}(-t_{\max}, t_{\max}), \quad
    s \sim \mathcal{U}(1-\Delta s,\, 1+\Delta s).
    \label{eq:sim2_sampling}
\end{equation}

Two transforms are applied per training pair. First, a \emph{common}, rotation-only transform
$T^{c}$, shared by both modalities, prevents the model from overfitting to the
axis-aligned walls that dominate floorplans. Second,
a \emph{noise} transform $T^{n}$, applied to a single modality, simulates the
residual misalignment; its
boundary parameters $(\theta^{n}_{\max}, t^{n}_{\max}, \Delta s^{n})$ are given
in \cref{sec:experiments} for training and in \cref{tab:difficulty_levels} for
the evaluation regimes.

\begin{figure}[t]
    \centering
    \includegraphics[width=\linewidth]{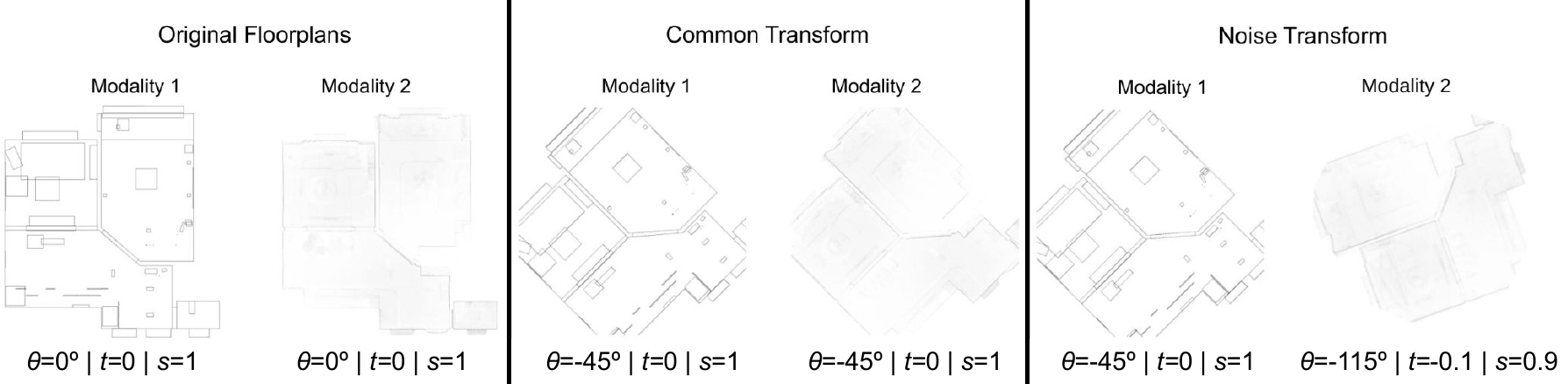}
    \caption{Training transformations. A common transform preserves alignment; a noise transform perturbs only one modality breaking cell-wise correspondences, which forces the model to learn meaningful representations instead of trivially representing cell indices. The inverse warp restores correspondence in latent space for supervision.}
    \label{fig:transforms}
\end{figure}

Since $T^{n}$ breaks the cell-wise correspondence between the two feature
grids, we do not compare $F$ and $F'$ directly. Instead, we warp the perturbed
grid back into the canonical frame using the known inverse flow field induced by
$T^{n}$. An example of these transformations can be seen in \cref{fig:transforms}.  

Note that this models residual misalignment as a global $\mathrm{Sim}(2)$
transformation. Real modality pairs may additionally differ through missing or
added structures, renovations, or non-rigid discrepancies between as-planned and
as-built geometry, which this family does not capture.

\section{Experiments}
\label{sec:experiments}
We evaluate local feature quality and downstream tasks in the held out validation set of Structured3D \cite{Structured3D}.
Inputs are resized to $256 \times 256$, using padding to keep the proportions of the floorplan; and MMFE outputs a $32\times32\times32$ grid.

\paragraph{Implementation details.}
The encoder uses a frozen DINOv3 ViT-B/16 backbone with a trainable MoGe-style DPT head that produces a $32\times32$ grid of $D=32$-dimensional descriptors; only the head is trained.
We optimize the per-cell InfoNCE objective of \cref{eq:cell_infonce} with temperature $\tau=0.07$, using AdamW ($\beta=(0.9,0.999)$, weight decay $10^{-4}$) at a constant learning rate of $10^{-4}$ (no scheduler), for $100$ epochs in mixed precision on a single NVIDIA RTX~4090 GPU.
Each mini-batch contains $32$ floorplan pairs, i.e. $64$ images, and the negatives for a cell are drawn only from the same mini-batch: they comprise all $64\times32\times32$ cells in the batch except the query cell and its cross-modal positive, i.e.\ $64\times32\times32-2 = 65{,}534$ negatives per cell.
Every training example pairs two randomly sampled modalities of the same floorplan, drawn from its available renderings, density maps, and the synthetic 2D-LiDAR point observation of \cref{sec:dataset}.
For the geometric-robustness recipe (\cref{fig:transforms}) we apply a \textit{common} rotation, $T^{c}$, of up to $\pm180^\circ$ to both modalities; and a noise similarity transform, $T^{n}$, to a single modality, with rotation in $\pm25^\circ$, translation in $\pm0.2$ of the image size, and scale in $[0.5,1.5]$.

For evaluation, we define three difficulty levels: \textit{easy}, \textit{medium} and \textit{hard}, based on the magnitude of the noise similarity transform, $T^{n}$. This is summarized in  \cref{tab:difficulty_levels}.

\begin{table}[ht]
\centering
\caption{Parameters defining the \textit{similarity} noise transformation $T^n$ difficulty levels.}
\setlength{\tabcolsep}{6pt}
\begin{tabular}{l c c c}
\hline
\textbf{Setting} & \textbf{Rotation ($\theta$)} & \textbf{Translation ($t$)} & \textbf{Scale ($\Delta s$)} \\
\hline
Easy   & 0°   & 0    & 0 \\
Medium & 15° & 0.1 & 0.2 \\
Hard   & 180° & 0.2 & 0.5 \\
\hline
\end{tabular}
\label{tab:difficulty_levels}
\end{table}

\subsection{Feature Quality and Modality Consistency}

We first evaluate the modality consistency of dense features at the cell level.
Given a pair of modalities for the same floorplan, we match each cell in one modality to its nearest neighbor in the other modality by cosine similarity and report (i) \textbf{Accuracy} (percentage of cells that retrieve the correct correspondence), (ii) Average End Point Error (\textbf{AEPE}) (average Euclidean error in grid coordinates), and (iii) Percentage of Correct Keypoints (\textbf{PCK@3}) (fraction of correspondences within three grid cells).
To justify our design, we compare MMFE (DINOv3 + DPT head) to two alternatives that use a CNN projection head on top of either a ResNet50 backbone~\cite{He2015DeepRL} or only the last-layer DINOv3 features~\cite{dinov3}.
As shown in \cref{tab:general_metrics_s3d}, the DINOv3 + DPT design, which also exploits intermediate backbone features, yields the most consistent cross-modal correspondences across difficulty levels.

\begin{table}[ht]
\centering
\small
\caption{Backbone/head comparison for cross-modal cell correspondence on Structured3D~\cite{Structured3D}, across difficulty levels.}
\setlength{\tabcolsep}{6pt}
\begin{tabular}{l l c c c}
\hline
Model & Difficulty & Acc ($\uparrow$) & AEPE ($\downarrow$) & PCK@3.0 ($\uparrow$) \\
\hline
\multirow{3}{*}{\makecell[l]{ResNet50 \\ + CNN}}
 & easy   & 0.18 & 15.01 & 17.63 \\
 & medium & 0.00 & 13.58 & 10.79 \\
 & hard   & 0.00 & 12.66 & 8.67 \\
\hline
\multirow{3}{*}{\makecell[l]{DINOv3 \\ + CNN}}
 & easy   & 29.22 & 1.30 & 91.93 \\
 & medium & 15.40 & 1.59 & 89.04 \\
 & hard   & 7.58 & 2.19 & 83.53 \\
\hline
\multirow{3}{*}{\makecell[l]{DINOv3 \\ + DPT\\  (Ours)}}
 & easy   & \textbf{46.00} & \textbf{0.78} & \textbf{94.53} \\
 & medium & \textbf{28.67} & \textbf{1.20} & \textbf{91.05} \\
 & hard   & \textbf{21.82} & \textbf{1.26} & \textbf{89.82} \\
\hline
\end{tabular}
\label{tab:general_metrics_s3d}
\end{table}

\begin{figure}[t]
    \centering
    \begin{subfigure}[b]{0.49\linewidth}
        \centering
        \includegraphics[width=\linewidth]{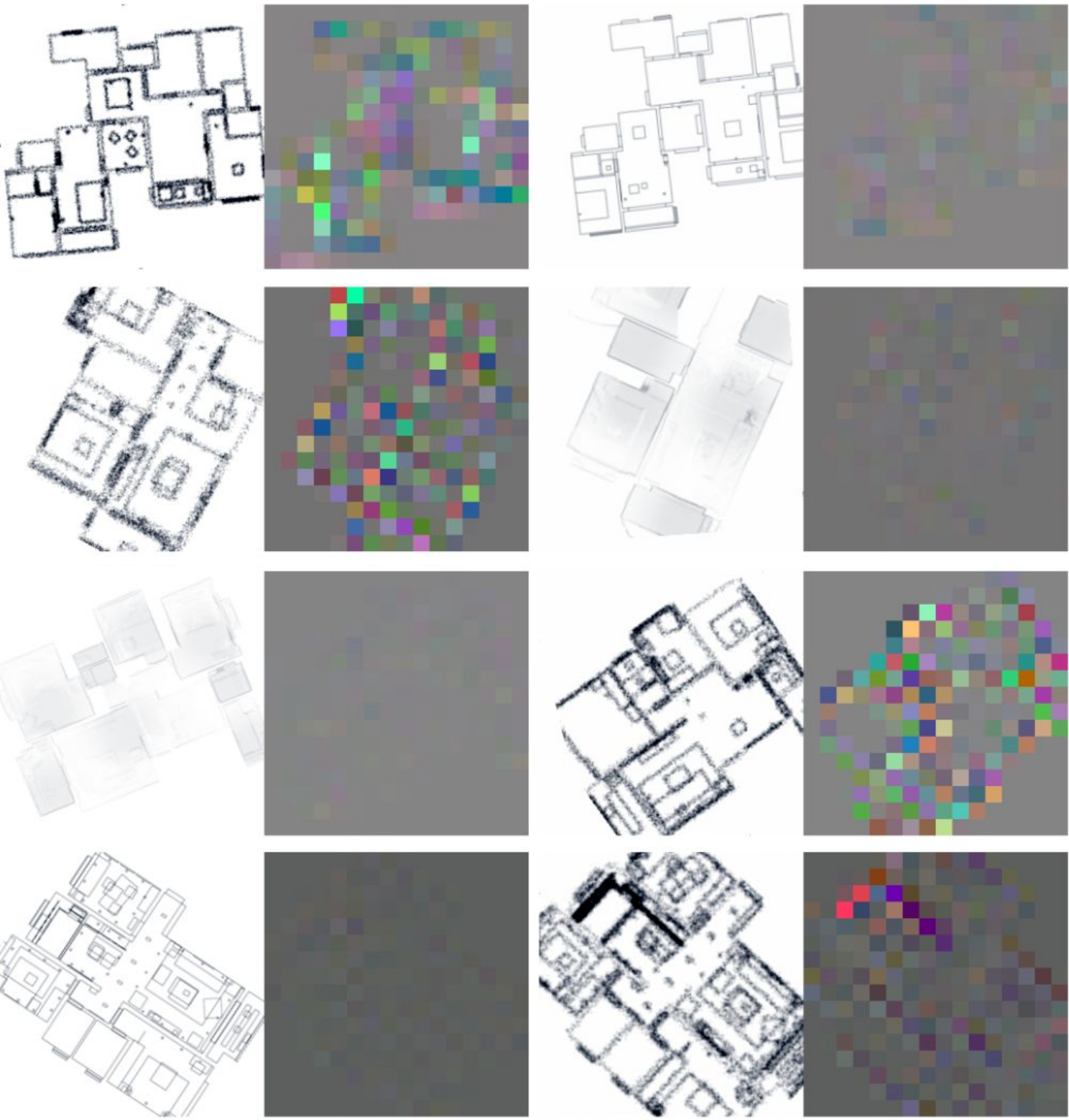}
        \caption{DINOv3 (PCA)}
        \label{fig:dino_pca}
    \end{subfigure}
    \hfill
    \begin{subfigure}[b]{0.49\linewidth}
        \centering
        \includegraphics[width=\linewidth]{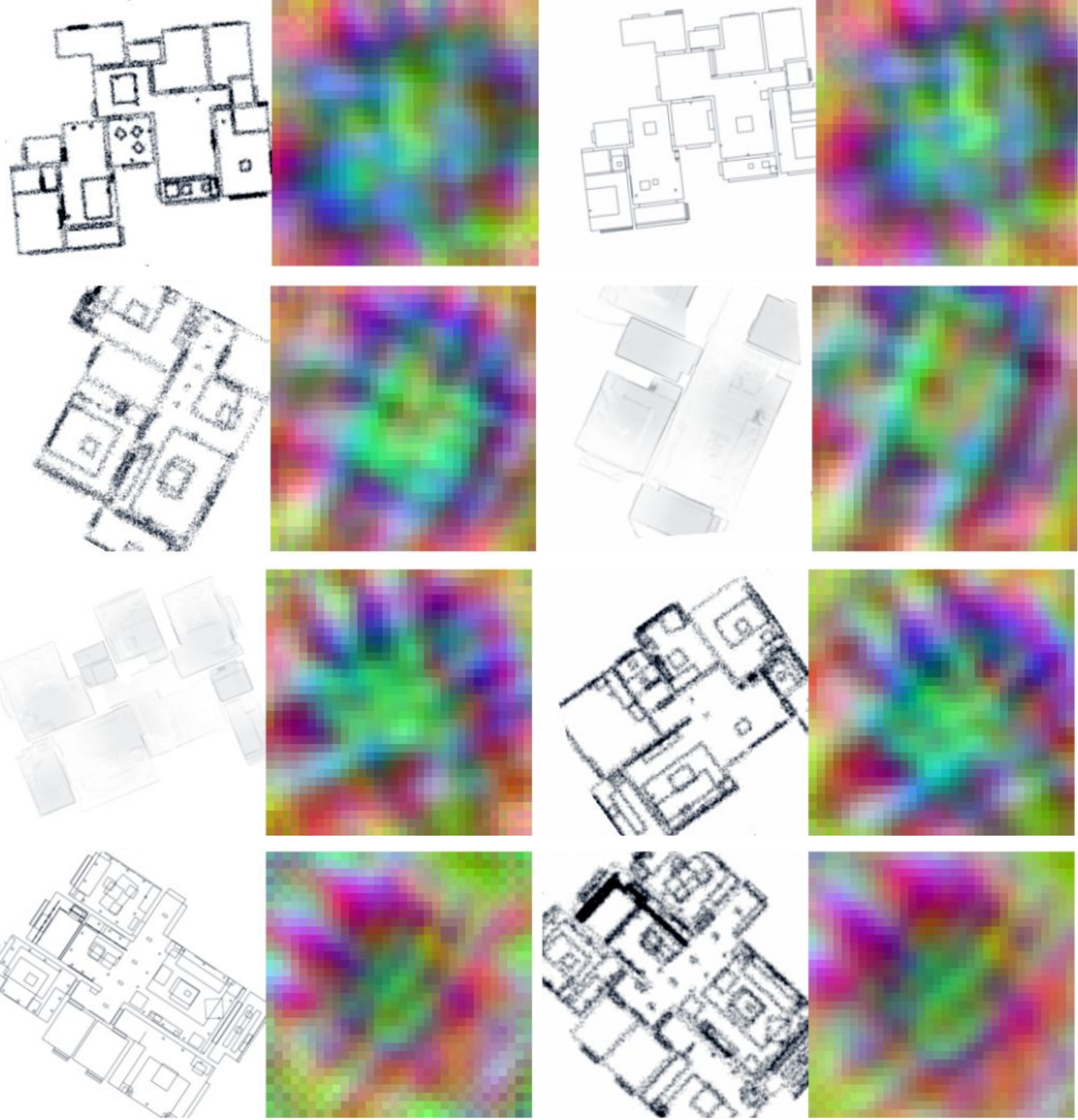}
        \caption{MMFE (PCA)}
        \label{fig:mmfe_pca}
    \end{subfigure}
    \caption{PCA visualization of dense features. DINOv3 is appearance-driven; MMFE aligns spatial structure across modalities.}
    \label{fig:pca}
\end{figure}

\Cref{fig:pca} visualizes PCA projections of dense feature grids produced by DINOv3 and by MMFE.
For the same floorplan across modalities, DINOv3 features vary substantially, reflecting appearance differences between modalities.
In contrast, MMFE produces highly consistent feature maps across modalities while still preserving meaningful spatial structure, even though the training objective only enforces local cross-modal correspondence.

\subsection{Cross-Modal Alignment}
\begin{figure}[t]
    \centering
    \includegraphics[width=0.8\linewidth]{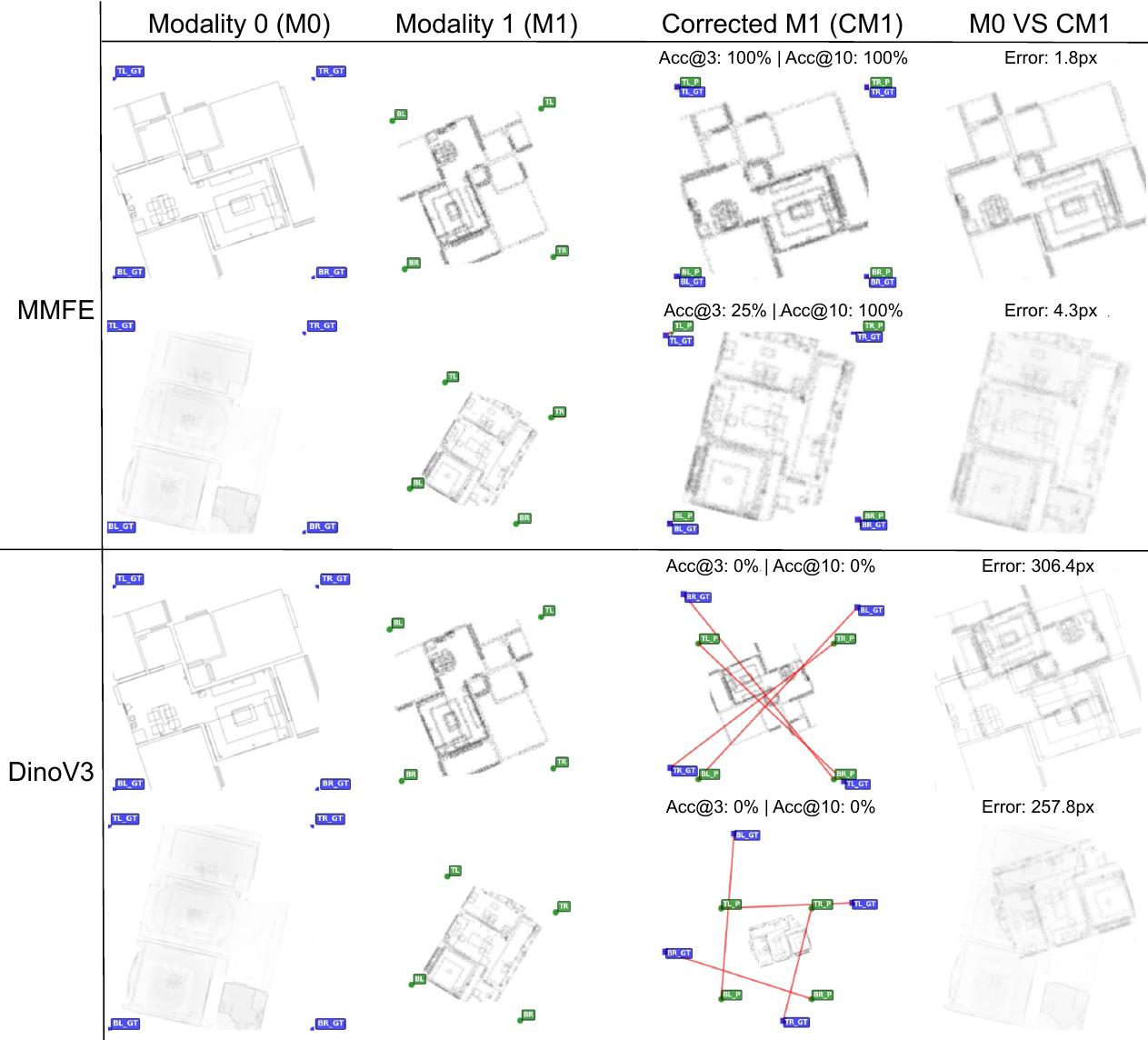}
    \caption{Qualitative alignment examples; red segments in the third column visualize residual corner error. Last column overlays the original M0 modality and the M1 modality after alignment is applied. 
    }
    \label{fig:align_qual}
\end{figure}

\paragraph{Problem statement.}
Let $x$ and $x'$ denote two observations of the same floorplan in different modalities, not necessarily spatially aligned: a pixel $(u,v)$ in $x$ does not need to correspond to the same physical location as the pixel $(u,v)$ in $x'$. We seek a planar similarity transformation $a \in \mathrm{Sim}(2)$ (rotation, translation, and scaling) that aligns $x'$ to $x$, formally $x(u,v) \approx x'\big(a(u,v)\big)$. \Cref{fig:align_qual} shows qualitative results of aligning two modalities with our method.

In the coarse setting, we estimate the $a\in \mathrm{Sim}(2)$ directly from the floorplan feature maps. Dense correspondences are obtained by nearest-neighbor matching in latent space and estimate a similarity transform with RANSAC~\cite{ransac}, using a partial-affine model with a minimal sample of two correspondences and a $3$-cell inlier (reprojection) threshold.
To handle rotations beyond the training range ($\pm25^{\circ}$), we rotate one modality in $45^\circ$ increments at test time and keep the hypothesis with the largest inlier count.This search is part of the alignment pipeline and not of the representation: it is applied identically to all methods in \cref{tab:alignment}, so the comparison against DINOv2/DINOv3 remains like-for-like.
Following standard practice~\cite{superpoint}, we map the four image corners with the estimated and the ground-truth transform and report, at the $256\times256$ input resolution, the RMS and median corner errors (RMSE and Med~RSE, in pixels) together with Acc@$k$ (the fraction of corners whose reprojection error is below $k$ pixels).

In \cref{tab:alignment}, MMFE substantially outperforms generic dense features (DINOv2~\cite{dinov2}/DINOv3 \cite{dinov3}) across both medium and hard settings, demonstrating the benefit of explicitly training for cross-modal geometric consistency.
We omit the \textit{easy} setting, where the inter-modality transform is the identity and alignment is trivially near-perfect. The accuracy nonetheless drops markedly between Acc@10 and Acc@5: for a $256\times256$ input we reduce the spatial resolution to $32\times32$, so each cell corresponds to an $8\times8$-pixel patch, which lower-bounds the attainable corner precision of the coarse matcher.

\begin{table}[t]
\centering
\caption{Alignment on Structured3D~\cite{Structured3D} via nearest-neighbor matching + RANSAC.}
\resizebox{\textwidth}{!}{
\begin{tabular}{l l c c c c c c c}
\hline
Model & Diff & RMSE($\downarrow$) & Median RSE($\downarrow$) & Acc@1($\uparrow$) & Acc@3($\uparrow$) & Acc@5($\uparrow$) & Acc@10($\uparrow$) & Acc@20($\uparrow$) \\
\hline
\multirow{2}{*}{\makecell[l]{DINOv2}} 
 & med  & 217.06 & 247.48 & 0.03 & 0.10 & 0.17 & 0.71 & 2.90 \\
 & hard & 239.34 & 261.57 & 0.00 & 0.07 & 0.07 & 0.24 & 0.84 \\
\hdashline
\multirow{2}{*}{\makecell[l]{DINOv3}} 
 & med  & 128.96 & 37.14 & 0.10 & 1.62 & 4.62 & 14.30 & 31.58 \\
 & hard & 168.32 & 124.91 & 0.03 & 0.84 & 2.02 & 5.13 & 13.25 \\
\hline
\multirow{2}{*}{\makecell[l]{MMFE (ours)}} 
 & med  & \textbf{11.70} &\textbf{ 4.41} & \textbf{7.09} & \textbf{36.91} & \textbf{58.87} & \textbf{84.38} & \textbf{92.98} \\
 & hard & \textbf{15.26} & \textbf{5.48} & \textbf{4.32} & \textbf{27.19} & \textbf{49.36} & \textbf{75.91} & \textbf{90.08} \\
\hline
\end{tabular}
}
\label{tab:alignment}
\end{table}

\paragraph{Fine-tuning a pretrained matcher.}
For sub-cell accuracy we pair MMFE with a state-of-the-art dense matcher, RoMa~\cite{roma,edstedt2025romav2}, on top of frozen backbone features.
We initialize from the released weights and fine-tune only the matcher and the
fine-level features, keeping the backbone (DINO or MMFE) frozen. Note that we finetune only RoMaV1, as the RoMaV2 training code was unavailable at the time of this work.

Results are given in \cref{tab:roma_mmfe_results}. First of all, comparing the first two rows and last two rows, it appears that a simple fine-tuning of the matching layers on top of our MMFE features is able to outperform the vanilla RoMa matching across modalities. However, the best results are obtained with DINO rather than MMFE features, despite the clear advantage of MMFE in coarse alignment.
Nevertheless, this is not a like-for-like comparison of feature quality: RoMa's released matcher was itself trained on top of frozen DINOv2 coarse features, so the DINOv2 rows benefit from a matcher already co-adapted to those features, while the MMFE rows require it to be re-adapted to a new feature space. Since RoMa's pretraining uses orders of magnitude more data than both MMFE's pretraining and our fine-tuning setup, fine-tuning the matching layers on our comparatively small dataset is insufficient for changing RoMa's matcher to work properly with our features. We disentangle the two factors below.

\begin{table*}[ht]
\centering
\caption{Fine-tuning a pretrained dense matcher (RoMa) with different backbone features, both out-of-the-box (OOB) and fine-tuned on our floorplan dataset (FT). Under this heavily pretrained regime, DINO features remain competitive with MMFE (see \cref{sec:limitations}).}
\resizebox{\textwidth}{!}{
\begin{tabular}{l l l c c c c c c}
\hline
Matcher & Features & Diff & RMSE($\downarrow$) & Med RSE($\downarrow$) & Acc@1($\uparrow$) & Acc@5($\uparrow$) & Acc@10($\uparrow$) \\
\hline
\multirow{2}{*}{RoMa v1 (OOB)}
& \multirow{2}{*}{DINOv2}
 & med  & 13.10 & 2.43 & 27.09 & 74.26 & 85.09  \\
&  & hard & 22.48 & 2.74 & 24.70 & 70.14 & 80.74 \\
\hdashline
\multirow{2}{*}{RoMa v2 (OOB)}
& \multirow{2}{*}{DINOv3}
 & med  & 31.26 & 2.59 & 30.57 & 71.19 & 80.70 \\
&  & hard & 54.72 & 2.58 & 28.27 & 67.95 & 76.86 \\
\hline

\multirow{2}{*}{RoMa v1 (FT)}
& \multirow{2}{*}{DINOv2}
 & med  & \textbf{8.66} & 2.05 & 35.80 & \textbf{82.73} & \textbf{89.51}  \\
&  & hard & \textbf{7.54} & \textbf{1.93} & \textbf{35.83} & \textbf{81.41} & \textbf{89.74} \\
\hline
\multirow{2}{*}{RoMa v1 (FT)}
& \multirow{2}{*}{MMFE}
 & med  & 11.12 & \textbf{2.04} & \textbf{37.82} & 79.93 & 87.15  \\
&  & hard & 13.35 & 1.98 & 34.51 & 77.87 & 86.47  \\
\hline
\end{tabular}
}
\label{tab:roma_mmfe_results}
\end{table*}

\paragraph{Isolating the contribution of the features.}
To disentangle feature quality from RoMa's pretraining advantage, we compare features under an equal data regime. We train RoMa from scratch on our floorplan dataset using either DINO or MMFE features, keeping the backbone frozen and learning only the matcher and the fine-level features.
As shown in \cref{tab:roma_mmfe_from_scratch}, MMFE features yield an average improvement of $20$ percentage points in accuracy over DINO, showing that the modality-invariance learned by MMFE directly benefits fine matching when the matcher is trained under matched conditions.

\begin{table*}[ht]
\centering
\caption{Training the RoMa matcher from scratch in the same data regime: with identical data and training, MMFE features improve accuracy by $20$ percentage points on average over DINO features.}
\resizebox{\textwidth}{!}{
\begin{tabular}{l l l c c c c c c}
\hline
Matcher & Features & Diff & RMSE($\downarrow$) & Med RSE($\downarrow$) & Acc@1($\uparrow$) & Acc@5($\uparrow$) & Acc@10($\uparrow$) \\
\hline
\multirow{2}{*}{RoMa w/o pretraining}
& \multirow{2}{*}{DINOv2}
 & med  & 35.52 & 7.79 & 12.28 & 42.85 & 57.35  \\
&  & hard & 47.10 & 11.42 & 8.74 & 38.12 & 48.92  \\
\hline
\multirow{2}{*}{RoMa w/o pretraining}
& \multirow{2}{*}{MMFE}
 & med  & \textbf{19.08} & \textbf{2.91} & \textbf{28.24} & \textbf{69.64} & \textbf{79.99}  \\
&  & hard & \textbf{23.71} & \textbf{3.18} & \textbf{24.83} & \textbf{61.61} & \textbf{73.89} \\
\hline
\end{tabular}
}
\label{tab:roma_mmfe_from_scratch}
\end{table*}

\subsection{Cross-Modal Retrieval}
\paragraph{Problem statement.}
Consider a database $\mathcal{X}=\{x_i\}_{i=1}^N$ of floorplans from multiple scenes and modalities, with $x'\in\mathcal{X}$. Given a query $x$, the objective is to retrieve the element of $\mathcal{X}$ corresponding to the same underlying scene but expressed in a different modality. Unlike in image retrieval, appearance is not a reliable cue: two observations of the same scene may be visually dissimilar because of modality, rendering style, noise or sampling density, so retrieval must instead rely on higher-level structural and semantic cues.

We aggregate local descriptors into global embeddings with two learnable
modules, NetVLAD~\cite{arandjelovic2016netvlad} and
SALAD~\cite{izquierdo2024optimal}, both trained (or fine-tuned) on our
multimodal floorplan training set with a global contrastive loss on paired
modalities, and report Top-$k$ accuracy. In \cref{tab:retrieval}, MMFE
consistently improves over generic DINO features for each aggregator under the
same training protocol, confirming that local multimodal alignment translates
into stronger global descriptors.

To assess the influence of the contrastive pretraining, we compare the exact same architecture (DINOv3\,+\,DPT\,+\,aggregator) trained end-to-end against a two-stage variant that first performs the contrastive pretraining and then trains only the
aggregator on frozen features. We keep the Dino features frozen, and we train the DPT + NetVLAD aggregator jointly with the goal of seeing the influence of the pretrained weights of the DPT head. As shown in \cref{tab:retrieval_pretraining}, the task-agnostic multimodal pre-training yields an approximately $38\%$ gain in Top-1 accuracy, showing that our multimodal pre-training can benefit specialized downstream setups.

\begin{figure}[t]
    \centering
    \includegraphics[width=0.9\linewidth]{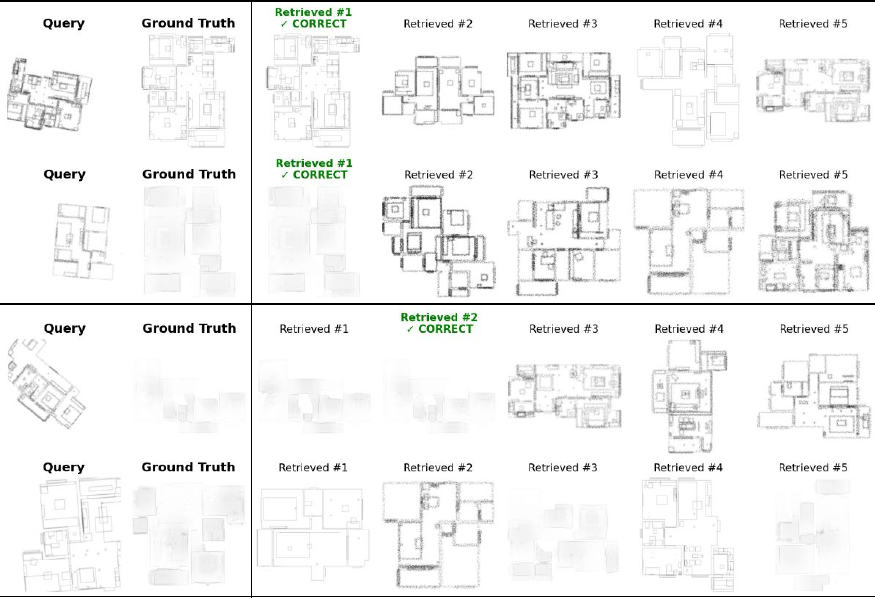}
    \caption{Qualitative top-5 retrieval examples on Structured3D~\cite{Structured3D} validation.}
    \label{fig:retrieval_qual}
\end{figure}

Qualitatively (\cref{fig:retrieval_qual}), even incorrect top-1 retrievals are structurally similar layouts, indicating that the embedding captures global geometric cues beyond modality-specific appearance.

\begin{table}[t]
    \centering
    \caption{Cross-modal floorplan retrieval on the Structured3D validation set. 
    }
    \setlength{\tabcolsep}{6pt}
    \begin{tabular}{l l c c c}
        \hline
        Aggregator & Encoder & Acc($\uparrow$) & Top5 Acc($\uparrow$) & Top10 Acc($\uparrow$) \\
        \hline
        \multirow{2}{*}{NetVLAD}
         & DINOv3  & 0.40 & 3.64 & 4.45 \\
         & MMFE & \textbf{24.29} & \textbf{40.08} & \textbf{50.61} \\
        \hline
        \multirow{2}{*}{SALAD}
         & DINOv3 & 43.72 & 64.77 & 72.06 \\
         & MMFE & \textbf{55.06} & \textbf{74.49} & \textbf{79.35} \\
        \hline
    \end{tabular}
    \label{tab:retrieval}
\end{table}

\begin{table}[ht]
\centering
\caption{Comparing the same architecture with and without contrastive pretraining. The task-agnostic pretraining improves Top-1 accuracy by approximately $38$ percentage points.}
\setlength{\tabcolsep}{6pt}
\begin{tabular}{l c c c}
\hline
Encoder & Acc($\uparrow$) & Top5 Acc($\uparrow$) & Top10 Acc($\uparrow$) \\
\hline
MMFE w/o Pretraining & 15.38 & 28.34 & 39.27 \\
\hline
MMFE w/ Pretraining & \textbf{53.04} & \textbf{70.85} & \textbf{76.52} \\
\hline
\end{tabular}
\label{tab:retrieval_pretraining}
\end{table}

\section{Limitations and Future Work}
\label{sec:limitations}

\paragraph{Data scale and realism.}
Our training set is small by the standards of deep learning: MMFE is trained on
roughly $45$K modality pairs, whereas dense matchers such as RoMa~\cite{roma} are
pretrained on over $2.5$M. It is also limited in realism on the sensing side: the
\textit{Sparse Range Scans} are simulated from the layout mask
(\cref{sec:dataset}) and the held-out Structured3D domain is itself synthetic, so
the generalization we report is across datasets and modalities rather than across
the real sensing process. Robustness to real range data, including noise,
occlusions, incomplete coverage, clutter, and the drift and non-metric distortion
of scanned maps, remains to be demonstrated and is the most important next step
for this work.

\paragraph{Resolution and scalability.}
MMFE outputs an $8\times8$-pixel patch per cell, which lower-bounds the corner precision attainable by the coarse matcher (\cref{tab:alignment}); \cref{sec:limitations} discusses how this bound could be raised.
This is a training-memory constraint rather than an architectural one: the DPT head can, in principle, produce finer grids, but the memory footprint of the per-cell InfoNCE loss grows with the number of cells; sampling a subset of cells during training would trade the number of negatives for memory.
Scaling from a single consumer-grade GPU to higher-end or multi-GPU setups would allow increasing both spatial resolution and descriptor dimensionality without reducing the number of negatives. 

\paragraph{Adapting large pretrained models to new features.}
Fine-tuning a large pretrained matcher (RoMa) on MMFE features did not surpass its DINO-based counterpart, despite the clear advantage of MMFE in coarse matching (\cref{tab:roma_mmfe_results}).
Fine-tuning strategies that exploit new backbone features while retaining pretrained knowledge could transfer the coarse-matching gains of MMFE to the fine matcher.

\paragraph{Beyond contrastive training.}
Our per-cell InfoNCE objective relies on large batches to provide informative negatives, which limits its scalability.
Self-supervised objectives that require no explicit negatives, in particular a joint-embedding predictive architecture (JEPA)~\cite{assran2023ijepa} using different modalities as alternative augmentations of the same scene, are a promising alternative for learning modality-invariant floorplan representations. 
Relaxing the one-to-one positive assignment, for instance through soft or
optimal-transport assignments, would additionally accommodate pairs that are not
in exact correspondence.

\paragraph{Incorporating perspective images.}
Working purely from 2D floorplans discards the 3D scene information carried by perspective images.
Aligning our multimodal latent space with perspective images, as done in SNAP~\cite{sarlin2023snap}, would embed both floorplans and perspective views in a shared representation, unlocking perspective image localization in multimodal floorplan settings.

\section{Conclusion}
\label{sec:conclusion}

We introduced MMFE, a dense multimodal floorplan encoder that maps heterogeneous 2D indoor representations into a shared latent grid by coupling a frozen DINOv3 backbone with a trainable DPT-style head, trained with a per-cell InfoNCE objective and a similarity-perturbation recipe that enforces geometric consistency through feature-grid warping. 

On the held-out Structured3D domain, the learned descriptors prove strongly modality-invariant. For coarse alignment, nearest-neighbor matching followed by RANSAC substantially outperforms generic DINOv2/DINOv3 features under both moderate and large geometric perturbations. 
When higher precision is required, MMFE is a drop-in front-end for a dense matcher: under a matched, in-domain data regime it improves a from-scratch RoMa matcher by roughly $20$ percentage points over DINO features, indicating that its benefit is most visible when downstream training data is limited rather than at the scale of large pretrained matchers. For retrieval, MMFE consistently improves over DINO features with both NetVLAD and SALAD, with the task-agnostic contrastive pretraining as the main driver. 

Together, these results show that a single dense, modality-invariant representation can support geometry-centric floorplan tasks across heterogeneous modalities.

\section*{Acknowledgements}
We thank Shaohui Liu for insightful discussions and feedback throughout this project. We are also grateful to Professor Marc Pollefeys, along with the entire Computer Vision and Geometry (CVG) group at ETH, for fostering a supportive environment where ambitious ideas and creativity can thrive.

\section*{Disclosure of Interests}
The authors have no competing interests to declare that are relevant to the
content of this article.

%
%
\clearpage
\bibliographystyle{splncs04}
\bibliography{main}
\end{document}